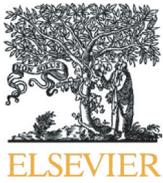
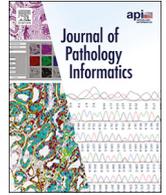

Review Article

# Performance of externally validated machine learning models based on histopathology images for the diagnosis, classification, prognosis, or treatment outcome prediction in female breast cancer: A systematic review

Ricardo Gonzalez [a,b,*,1], Peyman Nejat [c], Ashirbani Saha [d,e], Clinton J.V. Campbell [f,g], Andrew P. Norgan [h], Cynthia Lokker [i]

[a] *DeGroote School of Business, McMaster University, Hamilton, Ontario, Canada*
[b] *Division of Computational Pathology and Artificial Intelligence, Department of Laboratory Medicine and Pathology, Mayo Clinic, Rochester, MN, United States*
[c] *Department of Artificial Intelligence and Informatics, Mayo Clinic, Rochester, MN, United States*
[d] *Department of Oncology, Faculty of Health Sciences, McMaster University, Hamilton, Ontario, Canada*
[e] *Escarpment Cancer Research Institute, McMaster University and Hamilton Health Sciences, Hamilton, Ontario, Canada*
[f] *William Osler Health System, Brampton, Ontario, Canada*
[g] *Department of Pathology and Molecular Medicine, Faculty of Health Sciences, McMaster University, Hamilton, Ontario, Canada*
[h] *Department of Laboratory Medicine and Pathology, Mayo Clinic, Rochester, MN, United States*
[i] *Health Information Research Unit, Department of Health Research Methods, Evidence and Impact, McMaster University, Hamilton, Ontario, Canada*

ABSTRACT

Numerous machine learning (ML) models have been developed for breast cancer using various types of data. Successful external validation (EV) of ML models is important evidence of their generalizability. The aim of this systematic review was to assess the performance of externally validated ML models based on histopathology images for diagnosis, classification, prognosis, or treatment outcome prediction in female breast cancer. A systematic search of MEDLINE, EMBASE, CINAHL, IEEE, MICCAI, and SPIE conferences was performed for studies published between January 2010 and February 2022. The Prediction Model Risk of Bias Assessment Tool (PROBAST) was employed, and the results were narratively described. Of the 2011 non-duplicated citations, 8 journal articles and 2 conference proceedings met inclusion criteria. Three studies externally validated ML models for diagnosis, 4 for classification, 2 for prognosis, and 1 for both classification and prognosis. Most studies used Convolutional Neural Networks and one used logistic regression algorithms. For diagnostic/classification models, the most common performance metrics reported in the EV were accuracy and area under the curve, which were greater than 87% and 90%, respectively, using pathologists' annotations/diagnoses as ground truth. The hazard ratios in the EV of prognostic ML models were between 1.7 (95% CI, 1.2–2.6) and 1.8 (95% CI, 1.3–2.7) to predict distant disease-free survival; 1.91 (95% CI, 1.11–3.29) for recurrence, and between 0.09 (95% CI, 0.01–0.70) and 0.65 (95% CI, 0.43–0.98) for overall survival, using clinical data as ground truth. Despite EV being an important step before the clinical application of a ML model, it hasn't been performed routinely. The large variability in the training/validation datasets, methods, performance metrics, and reported information limited the comparison of the models and the analysis of their results. Increasing the availability of validation datasets and implementing standardized methods and reporting protocols may facilitate future analyses.

## Contents

\* Corresponding author at: DeGroote School of Business, McMaster University, 1280 Main Street West, Hamilton, Ontario L8S4M4, Canada
   *E-mail addresses:* rgonzalez@lunenfeld.ca, gonzar5@mcmaster.ca (R. Gonzalez).
[1] Present address: Lunenfeld-Tanenbaum Research Institute, Mount Sinai Hospital, Joseph & Wolf Lebovic Health Complex, 881–600 University Avenue, Toronto, Ontario M5G 1X5, Canada.

## Introduction

*Rationale*

Female breast cancer is the most commonly diagnosed cancer and, in women, the most frequent cause of cancer mortality.[1] Histopathological examination of breast tissue samples is the reference standard for cancer diagnosis and is used to determine the prognosis of a patient and risk factors to predict outcomes.[2,3]

Pathologists have been using microscopes with glass slides containing preserved human tissue to help make a diagnosis for more than 100 years. Now, these slides can be scanned and digitized to be viewed on computer screens.[4,5] The process of scanning glass slides to produce digital images (Whole-Slide Images or WSI) was initially called "Digital Pathology" (DP).[4] However, this term has evolved and is now used to encompass many other related processes in the modern pathology workflow.[4,5] DP has been approved and implemented in laboratories for routine diagnosis in many countries and is expected to increase in the near future.[6,7] Beyond other potential benefits, DP can facilitate the implementation of machine learning-based computational pathology tools to support diagnostic pathology workflows.[8]

ML model basic development steps include problem formulation,[9] ML algorithm selection,[10] data preparation,[10,11] ML model training,[12] hyperparameter tuning,[12] model evaluation,[12] and model deployment/ maintenance.[9,10] In a data-rich situation, the best approach to build a prediction/classification model is to randomly divide an input dataset into 3 parts: a training set, a validation set, and a testing set.[11,13] The former is used to fit the model, the second to estimate the prediction error for model selection, and the later to assess the generalization error of the selected model.[13] As the terminology employed to refer to ML models' evaluation and validation datasets could confuse readers,[12,14] the following terms are used here:

- Internal validation: Model evaluation conducted with data extracted from an input dataset (i.e., the "testing set" mentioned above)[11,15–17] that was set aside from the training/tuning dataset at the beginning of the study to evaluate the final version of a ML model a single time.[11]
- EV: Model evaluation conducted with data extracted from independent datasets[11,15–17] to evaluate the final version of a ML model a single time.[11]
- Training/Tuning datasets: Used for model training, optimization, and/or model selection.[11]

During tissue processing, slide preparation, slide digitization, image compression, and image storage, hidden variables (i.e., image data unrelated to the actual prediction/classification task that can affect the performance of ML models) may be introduced to WSIs.[18–20] As protocols, equipment, and consumables utilized during these processes vary among different institutions,[21] the independent datasets used during EV should ideally be extracted from a different data source, such as another clinic or hospital system.[11,15–17]

Only EV is considered important evidence of generalizability as patterns learned from hidden variables of training datasets (instead or in addition to those that can be learned from the intended target variables) are not expected to improve ML models' performance when they are tested with independent datasets.[11,15,18] Despite this, most of these models have not been externally validated.[16,17,22–28]

*Objective*

This systematic review aimed to assess the performance of externally validated ML models based on histopathology images for diagnosis, classification, prognosis, or treatment outcome prediction in female breast cancer.

**Material and methods**

*Eligibility criteria*

We conducted a systematic review following the PRISMA 2020 guidelines.[29] Studies focused on female breast cancer (invasive tumors or carcinomas in situ) that externally validated the performance of ML models using data extracted from histopathology images (stained with hematoxylin/eosin or other histochemical stains) directly for diagnosis, classification, prognosis, or treatment response prediction were included. The EV had to be suggested, implied, or mentioned in the Title/Abstract and conducted with breast cancer images for a study to be incorporated. In addition to those not meeting the above criteria, studies validating ML models developed to predict biomarkers were excluded.

*Information sources*

A systematic search of original research journal papers and conference proceedings written in English and published from January 1, 2010 to February 28, 2022 was conducted in March 2022 using the following databases: MEDLINE (via OVID), EMBASE (via OVID), CINAHL (via EBSCO), IEEE (via IEEE Xplore), MICCAI (via Springer link), and SPIE conferences. A brief description of these databases is presented in Appendix A. The search strategies were developed by authors and reviewed by a health sciences librarian with expertise in systematic reviews.

*Search strategy*

The search strategies are shown in Appendix B.

*Selection process*

Search results were imported into Rayyan[30] to remove duplicates and for title/abstract screening. Two reviewers independently conducted the initial title and abstract screening. Selected full-text articles were downloaded for all titles that met the inclusion criteria or where there was any uncertainty whether the inclusion criteria were met (e.g., when it was unclear if EV was performed). Two reviewers independently conducted the full-text screening using Mendeley.[31] Additional information was requested from study authors when needed. A third person acted as an adjudicator to resolve any conflicts.

*Data collection process*

One author independently extracted data from included studies with verification by a second.

*Data items*

The following information was extracted: Authors, publication date, country of study, objectives, and main resus, including all performance measures in the EV of the ML models. In addition, related to the ML models, the following data were extracted: Algorithms employed; source, number, and type of images used for model development (training/tuning and internal validation datasets) and EV; histological type of tumors/entities contained in the training and EV datasets; details on preprocessing of the images; software platform(s) used for annotation/ground-truth preparation and computational purposes.

*Study risk of bias assessment*

The risk of bias by using PROBAST (prediction model risk of bias assessment tool) for non-randomized studies was assessed by one author and confirmed by another.

*Effect measures*

The effect measures were all performance metrics used to evaluate the ML model during EV, such as accuracy, area under the curve (AUC), precision, recall, etc.

*Synthesis methods*

The main study characteristics and findings are presented in the text and summarized in tables. Results are narratively described. A meta-analysis was not conducted because of the heterogeneity of the studies in algorithms, type of images, and reported outcomes.

**Results**

*Study selection*

The search queries identified 2157 articles and 182 conference proceedings. After removing 328 duplicates and excluding 1961 publications during the title and abstract screening, 50 were assessed during full-text screening, and 10 were included in the review (Fig. 1).

The number of records reviewed during the Title/Abstract screening progressively increased from 2010 to 2015, declined from 2015 to 2017, and rose again from 2017 to 2021. In 2022, only the records published from January 1 to February 28 are shown (Fig. 2).

Examples of records that passed and did not pass the Title/Abstract screening are provided in Appendix C.

*Study characteristics*

In the included studies, 3 evaluated ML models for diagnostic purposes,[32–34] 4 for classification purposes,[35–38] 2 for prognosis purposes,[39,40] and 1 for prognostic and classification purposes[41] were externally validated. Eight were reported in journal articles[33,34,36–41] and 2 in conference proceedings.[32,35] The datasets, preprocessing steps, and software used for annotation purposes are shown in Table 1.

The reported information about the type of images, and the software platforms used for annotating and computational purposes and the type of tissues and tumors were largely heterogeneous among different studies. In those where relevant information was available, WSIs were employed[32–34,36–41] and only 2 studies included TMA images.[39,40] Formalin-fixed, paraffin-embedded tissue samples were utilized in 3 studies,[37,40,41] and frozen section samples were used in 1 (for the internal validation dataset, in addition to paraffin-embedded tissues).[36] All tissues were stained with hematoxylin and eosin[32–34,36–41] and the most common scanning magnification was 40×.[33,34,36,38,41] The patches sizes ranged from 50 × 50[32] to 2048 × 1536 pixels.[36] The *procedures* employed to *remove the* breast tissues (e.g., core needle biopsies or mastectomies) were not specified. However, Radiya et al. stated that scanned images of "breast biopsies" were used.[37] Two authors reported the software platforms used for computational purposes. Bychkov et al. used PyTorch to implement deep learning architectures[40] and Wang et al. used Keras (2.2.4) framework with TensorFlow (1.12) backend.[41] Both used ADAM for optimization.[40,41]

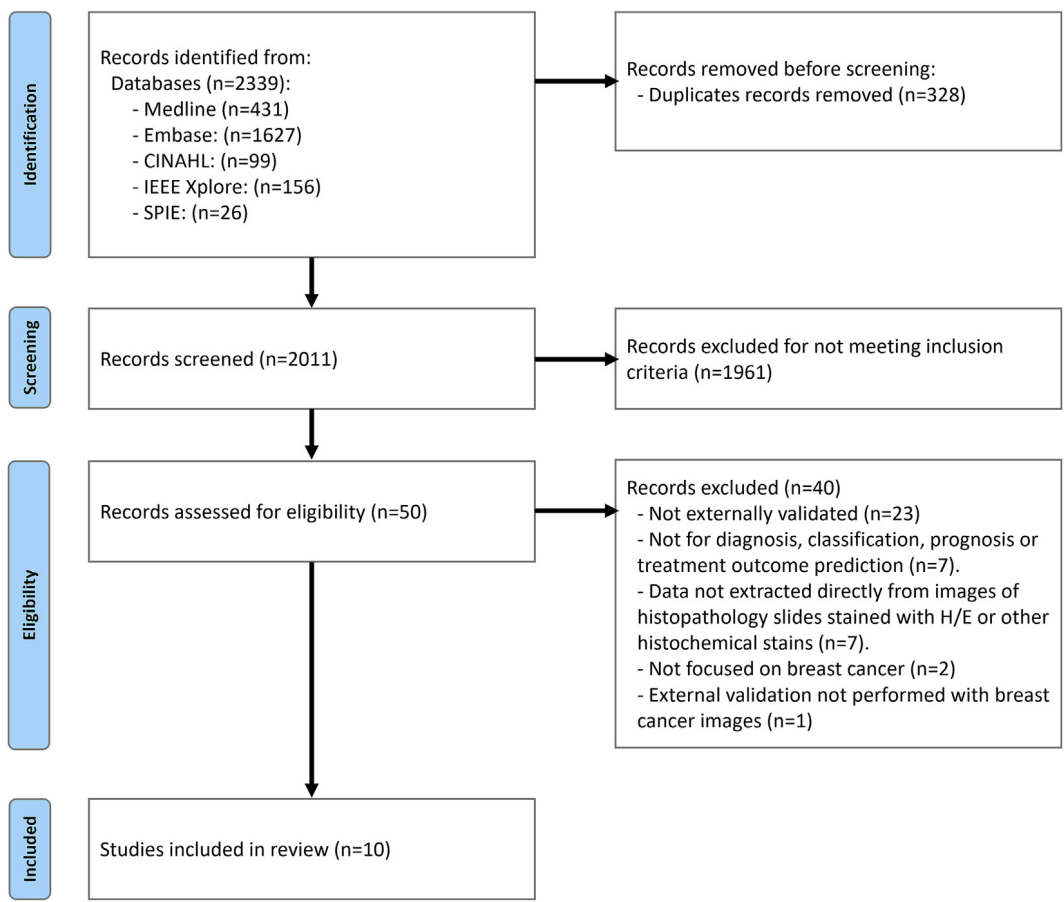

Fig. 1. PRISMA flow diagram of the studies identification process for the systematic review.

Regarding the algorithms, 9 studies used convolutional neural networks (CNN). ResNet[35,38,40] and DenseNet-161[38] were used as backbone networks in three studies and InceptionV3 in 2.[36,41] One study used a "Combined model with Active Feature Extraction" (CAFE), based on two logistic regression algorithms.[37] A detailed description of each algorithm is shown in Appendix D.

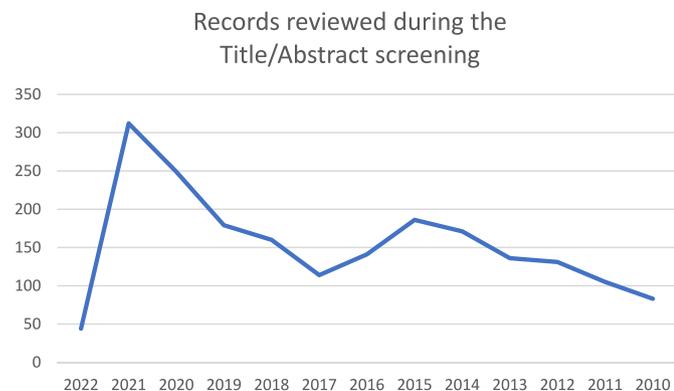

Fig. 2. Distribution of the records reviewed during the Title/Abstract screening by year (January 1, 2010–February 28, 2022).

### Risk of bias in studies

According to the PROBAST,[42] 2 studies were at a high risk of bias (ROB) in Domain 4 (Analysis) and the Overall judgment due to a small sample size of the validation dataset.[35,37] For other studies, the ROB in Domain 4 (Analysis) and the Overall ROB was unclear due to the lack of more detailed information about the statistical analyses (studies validating prognostic ML models),[39–41] the ROB in Domain 1 (Participants) and the Overall judgment was unclear due to the lack of more detailed information about the datasets (in all but one),[32–39,41] and the ROB in Domain 2 (Predictors) was unclear due to the lack of more detailed information on how predictors were defined or assessed.[35,37] There was "low concern" regarding applicability for all studies (i.e., the population, predictors, or outcomes of the studies matched the systematic review question)[32–41] (Fig. 3).

### Results of individual studies

The ground truth used to assess the performance of models developed for diagnostic (i.e., to detect breast cancer) or classification (i.e., to separate different lesions of the breast) purposes were pathologists' annotations/opinions and, for prognostic models (i.e., to predict survival of patients), clinical data. The results of the external validation are summarized below. Related internal validation results are presented in Appendix E.

### ML for diagnostic purposes

While Cano et al. selected accuracy as the only performance metric for the EV of their model,[32] Cruz Roa et al. reported several others.[33,34] During the EV, their models were employed to detect the presence or absence of breast cancer, separating cases with and without invasive ductal

Table 1

Datasets, sources, and preprocessing steps for development and validation of ML models and software used for annotation purposes.

| | Diagnostic models | | | Classification [continues] | |
|---|---|---|---|---|---|
| Study | Cano et al. (2018) | Cruz-Roa et al. (2017) | Cruz-Roa et al. (2018) | Colon-Cartagena et al. (2020) | Mi et al. (2021) |
| Country of study | Colombia & United States | Colombia & United States | Colombia & United States | United States | China |
| Type of tumors/entities | Invasive ductal carcinomas | ER + invasive breast cancer | ER + invasive breast cancer | High-grade DCIS | Training/IV: Normal, benign, DCIS, and invasive carcinomas EV: Entities from BreakHis[a] and BACH[b] |
| Training/tuning datasets: source (*n*) | HUP[c] (239) TCGA[c] (172) | HUP (239) CWRU/UHCMC (110) | HUP (239) CWRU (110) | VCU (334) | PUMCH (371) |
| IV datasets: source (n) | TCGA[c] (172) HUP[c] (239) | CINJ (40) | CINJ (40) | VCU (80) | PUMCH[d] (169) |
| EV datasets: source (n) | CINJ (40) | TCGA (195) CWRU/UHCMC (21)[e] | TCGA (195) | NS (31) | BreakHis (7909) BACH (430) |
| Preprocessing steps | NS | Color normalization Data augmentation | Color normalization Data augmentation | NS | Data augmentation Image resizing Random color perturbations |
| Software used for annotation | NS | ImageScope v11.2 (Aperio) Image Viewer v3.1.4 (Ventana) | ImageScope v11.2 (Aperio) Image Viewer v3.1.4 (Ventana) | NS | ASAP |

| | Classification [continued] | | Prognosis | | Prognosis & Classification |
|---|---|---|---|---|---|
| Study | Radiya-Dixit et al. (2017) | Yang et al. (2019) | Bai et al. (2021) | Bychkov et al. (2022) | Wang et al. (2021) |
| Country of study | United States | China | Sweden & United States | Finland | Sweden |
| Type of tumors / entities | DCIS and UDH | Training/IV: Entities from BACH[b] EV: Entities from BreakHis[a] | Triple-negative breast cancer | Breast cancer | Primary invasive breast cancer (types not specified) |
| Training / tuning datasets: source (*n*) | MGH (116)[f] | BACH (400) | Yale School of Medicine (95)[g] | FinProg (693) | SSGH, KUH, and TCGA (844) |
| IV datasets: source (n) | MGH (116) BIDMC (51)[f] | BACH (100) | Yale School of Medicine (171)[g] | FinProg (354) | SSGH, KUH, and TCGA (351) |
| EV datasets: source (n) | BIDMC (51)[f] | BreakHis (1995) | Yale School of Medicine (417)[g] WTS Sweden (216) TCGA (116) | FinHer trial (712) | SCAN-B (1262) |
| Preprocessing steps | NS | Data augmentation | NS | Color normalization Data augmentation | Color normalization Data augmentation |
| Software used for annotation purposes | Fiji (ImageJ, National Institutes of Health) | NS | QuPath | WebMicroscope | NS |

BIDMC: Beth Israel Deaconess Medical Center. CINJ: New Jersey Cancer Institute. CWRU: Case Western Reserve University. DCIS: Ductal Carcinoma in situ. ER +: Estrogen receptor-positive. EV: External validation. HUP: Hospital of the University of Pennsylvania. IV: Internal validation. KUH: Karolinska University Hospital. MGH: Massachusetts General Hospital. NS: Not specified. PUMCH: Peking Union Medical College Hospital. SCAN-B: Sweden Cancerome Analysis Network-Breast project. SSGH: Stockholm South General Hospital. TCGA: The Cancer Genome Atlas. UDH: Usual Ductal Hyperplasia. UHCMC: University Hospitals Case Medical Center. VCU: Virginia Commonwealth University.

[a] BreakHis dataset: (1) Benign lesions: Adenosis, Fibroadenomas, Phyllodes tumors, and Tubular adenoma. (2) Malignant tumors: Ductal carcinoma, Lobular carcinoma, Mucinous carcinoma, and Papillary carcinoma.

[b] BACH dataset: Normal, benign, carcinomas in situ, and invasive carcinomas.

[c] Internal validation with cases from the TCGA was conducted after the models were trained with cases from HUP, and internal validation with cases from HUP was conducted after the models were trained with cases from the TCGA.

[d] Internal validation dataset used by Mi et al. (2021) contained 115 paraffin-embedded tissues and 54 frozen section samples.

[e] External validation dataset used by Cruz-Roa et al. (2017) included positive and negative controls. Positive controls were extracted from the TCGA. Negative controls were extracted from normal breast tissue regions adjacent to invasive ductal carcinomas of patients diagnosed at UHCMC/CWRU.

[f] External validation performed with cases from the BIDMC after training the model with cases from the MGH. All cases from both institutions were combined to train and test the model when conducting the internal validation (using cross-validation).

[g] Training, IV, and EV datasets contained cases from different cohorts of patients of the Yale School of Medicine.

carcinoma[32] or with and without estrogen receptor-positive invasive breast cancer[33,34] (as a binary classification problem) (Table 2).

*ML for classification purposes*

The most common performance metric used in their EV was accuracy,[35,36,38] followed by area under the curve (AUC).[37,38] Mi et al. reported the accuracies of their model with 4 different magnifications when differentiating benign lesions (Adenosis, Fibroadenomas, Phyllodes tumors, and Tubular adenoma) from malignant tumors (Ductal carcinoma, Lobular carcinoma, Mucinous carcinoma, and Papillary carcinoma) in the BreakHis dataset (as a binary classification problem), and when separating cases labeled as normal, benign, ductal carcinoma in situ (DCIS), and invasive carcinoma in the BACH dataset, part A (as a multi-label classification problem).[36] Radiya-Dixit et al. reported the AUC of the ROC curve when their model was used to categorize breast lesions as either benign usual ductal hyperplasia or DCIS (a binary classification problem).[37] Yang et al. included the performance metrics for the entire dataset (overall accuracy, AUC, precision, and recall) and for the benign lesions and malignant tumors categories of the BreakHis dataset (as binary classification problems).[38] In summary, accuracy ranged from 87.2% (in the BACH dataset, part A)[36] to 99.75% (in the BreakHis dataset),[38] AUC from 91.8% (differentiating usual ductal hyperplasia from DCIS)[37] to 99.99% (BreakHis dataset),[38] and

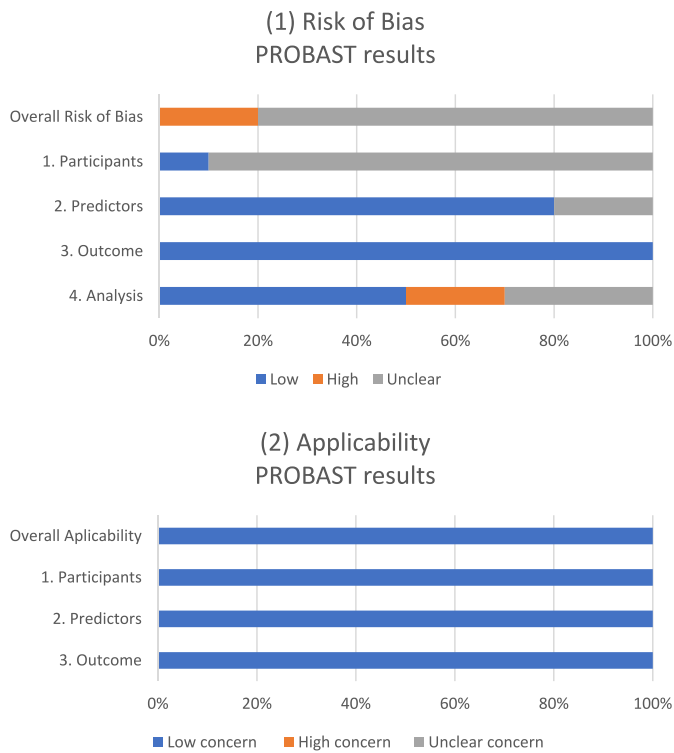

**Fig. 3.** Prediction model Risk Of Bias Assessment Tool (PROBAST) Graphical presentation—(1) risk of bias results and (2) applicability.

precision and recall from 99.20% to 100% (BreakHis dataset).[38] (Table 3)

*ML for prognostic purposes*

Bai et al. reported that all the tumor-infiltrating lymphocytes (TILs) variables had a significant prognostic association with overall survival ($P \leq .01$ for all comparisons). However, as shown in Table 4, the prognostic association of each TILs variable was different on each external validation dataset. In addition, the derived easTILs variable score had a good correlation with the pathologist-read sTILs in the WTS Sweden cohort (Spearman $r$ coefficient = 0.63, $P < .0001$).[39] Bychkov et al. found that both "Solo" models (i.e., to predict the distant disease-free survival data only) and those trained in a multitask fashion (i.e., predicting estrogen receptor and HER2 status together with the distant disease-free survival data) significantly predicted distant disease-free survival[40] (Table 4).

**Table 2**
Results of EV of diagnostic models.

| Study | Cano et al. (2018) | Cruz Roa et al. (2017) | | Cruz-Roa et al. (2018) |
|---|---|---|---|---|
| Dataset | | TCGA | CWRU/UHCMC[a] | |
| Accuracy (%) | 88.78 | – | – | – |
| Dice coefficient(%) | – | 75.86 | – | 76 |
| Positive-predictive value (%) | – | 71.62 | – | 72 |
| Negative-predictive value (%) | – | 96.77 | 100 | 97 |
| True-positive rate (%) | – | 86.91 | – | 87 |
| True-negative rate (%) | – | 92.18 | 99.64 | 92 |
| False-positive rate (%) | – | 7.82 | 0.36 | 8 |
| False-negative rate (%) | – | 13.09 | – | 13 |
| Heatmaps | – | – | – | √ |

CWRU: Case Western Reserve University. TCGA: The Cancer Genome Atlas. UHCMC: University Hospitals Case Medical Center. √: Good concordance between predictions of HASHI and pathologists' annotations.
[a] Not all performance metrics were calculated because the "normal" dataset did not have cancer annotations.

*ML for classification and prognosis purposes*

Wang et al. achieved an AUC of 0.907 (95% CI, 0.88–0.93, $P = .930$) when separating Nottingham Histological Grade 1 and Nottingham Histological Grade 3 invasive breast carcinomas and predicted recurrence-free survival rates between DeepGrade-classified Nottingham Histological Grade 1 and 3 patients that were similar to that of clinically assigned Nottingham Histological Grade 1 and 3 (defining recurrence as "locoregional or distant relapses, contralateral tumors or death"). It also provided a significant prognostic value for stratification of Nottingham Histological Grade 2 ($P = .0045$), and those predicted as high grade showed an increased risk for recurrence, with a HR of 1.91 (95% CI, 1.11–3.29, $P = .019$). Pathologists' opinions were used as the ground truth.[41]

**Discussion**

*Interpretation and implications of the results*

To our knowledge, this is the first systematic review specifically focused on assessing externally validated ML models based on histopathology images for diagnosis, classification, prognosis, or treatment outcome prediction in female breast cancer.

Validating ML models is essential to ensure they will perform the task they were developed for.[14] In an internal validation, an input dataset is split into parts; one is used to train and potentially fine-tune a ML model, and the other to test it. An EV uses independently derived datasets (i.e., external) to train and test ML models.[11,15–17] As patterns learned from hidden variables of training datasets are not expected to improve ML models' performance when tested with independent datasets, only EV is regarded as proof of generalizability.[11,15,18]

Although 2011 journal articles and conference proceedings were identified with our search queries, the majority of them did not report an EV of their ML models. This limitation has also been described for other ML models developed for medical purposes[16,17,22–28,43–45] and for diagnostic or predictive purposes on breast cancer patients specifically.[46–48] As previously stated, this may be related to the difficulty in finding appropriate external datasets,[47,48] non-adherence to guidelines that promote EV[11,46,48–51] and lack of awareness of their importance.[48] Particularly, DP is not yet widely adopted, and robust datasets with reliable labels and appropriate follow-up information have only become available only until recently.[20,52–56] Also, as some hold sensitive information or belong to private companies that cannot or do not want to share their data, they have not been publicly available.[47] This will possibly change with an increased awareness of DP applications and benefits and its widespread adoption in upcoming years.[55,57,58]

The reason why more of the included studies validated ML models for diagnosis or classification purposes is unknown. However, it may be partly explained by the fact that ML models used to predict biomarkers were excluded during the selection process. This could have limited the number of eligible studies used for prognosis and prevented the inclusion of any utilized for treatment-response prediction.

The commercial introduction of scanners that could generate high-resolution WSIs began in the 1990s.[59,60] However, as described by Cooper et al., only during the past half-decade ML models have become a driving force for advancements in pathology.[60] This work encompasses the time when this domain gained and sustained traction and, similarly, based on the records reviewed during the Title/Abstract screening, the number of publications has been progressively increasing in the last half-decade.

Limited access to large datasets to train ML models may explain why most studies were conducted in the United States/Colombia, the United States, or China.[61] Although Colombia has not been traditionally considered as a leader in artificial intelligence (AI)-related research output, in 3 studies, the first authors were affiliated with Colombian institutions and used datasets extracted from the United States.

There was significant heterogeneity in the studies regarding the amount and type of information reported. In addition, the performance metrics,

Table 3

Results of EV of classification models.

| Study | Colon-Cartagena et al. (2020) | Mi et al. (2021) | | | Radiya-Dixit et al. (2017) | Yang et al. (2019) | |
|---|---|---|---|---|---|---|---|
| Dataset/category | | BreakHis (Magnification) | BACH, part A | BACH, part B | | Benign | Malignant |
| Accuracy (%) | 90 | 96.7 (4×) | 87.2 | – | – | 99.75 | 99.75 |
| | | 97.6 (10×) | | | | | |
| | | 95.0 (20×) | | | | | |
| | | 93.3 (40×) | | | | | |
| Area under the curve (%) | – | – | – | – | 91.8 | 99.99 | 99.99 |
| Precision (%) | – | – | – | – | – | 100 | 99.2 |
| Recall (%) | – | – | – | – | – | 99.64 | 100 |
| Heatmaps | – | – | – | √ | – | – | – |

√: Good concordance between predictions of HASHI and pathologists' annotations.

datasets, ML models, image preprocessing, patch sizes, magnifications, and platforms used for annotating/computational purposes were highly variable. This finding corroborates the lack of standardization on the methodology and reported information described by Fell et al.[62] and found in previous systematic reviews, such as those reported by Mazo et al. with studies using AI tools to predict breast cancer recurrence,[47] by Gao et al. with ML-based breast cancer risk prediction models,[63] by Corti et al. with AI algorithms for prediction of treatment outcomes in breast cancer,[46] by Nagendran et al. with deep learning algorithms for medical imaging,[16] and by Yu et al. with deep learning algorithms with EV for radiologic diagnosis.[48] As explained by other authors, this is a noteworthy limitation of these systematic reviews[48] that impedes from making rigorous comparisons,[63] better understand findings,[64] and limits models' generalizability and their clinical impact.[46] Increasing adherence to existing and upcoming reporting guidelines[11,26,49,50,65,66] could be potentially improved by training authors on their practical use, enhancing the understanding or their content, encouraging and checking the adherence to them, and involving experts on methodology and reporting on AI research groups.[67]

It is recommended[14] to test ML models with diverse and large datasets that address the variability found in real-world data and allow statistically meaningful analysis. Nevertheless, it was impossible to determine if the included studies were aligned with this recommendation. That is because detailed descriptions of the variability of the datasets were not included in the reports. In addition, as stated above, important differences among the studies in terms of the source, size, and histological type of tumors/entities included in the training/tuning and validating datasets were noted.

The majority of experiments were performed using WSIs. As for many other tools developed for computational pathology,[68] most authors utilized

Table 4

Results of EV of prognostic models for predicting survival.

| Study | Bai et al. (2021) | | | | Bychkov et al. (2022) | |
|---|---|---|---|---|---|---|
| Dataset/Model | TMA Yale1 | TMA Yale2 | WTS TCGA | WTS Sweden | "Solo" model | Multitask model |
| | Hazard ratios[a] (95% CI) *P*-value | | | | | |
| High eTILs% | 0.64 | 0.43 | 0.09 | NS | – | – |
| | (0.43–0.94) | (0.26–0.69) | (0.01–0.70) | | | |
| | 0.025 | 0.0005 | 0.02 | | | |
| High etTILS% | 0.51 | 0.47 | 0.1 | NS | – | – |
| | (0.32–0.81) | (0.28–0.77) | (0.01–0.80) | | | |
| | 0.004 | 0.003 | 0.03 | | | |
| High esTILs | 0.48 | 0.42 | NS | NS | – | – |
| | (0.25–0.89) | (0.24–0.76) | | | | |
| | 0.02 | 0.004 | | | | |
| High eaTILs (mm$^2$) | 0.48 | 0.62 | 0.1 | NS | – | – |
| | (0.31–0.74) | (0.37–1.01) | (0.01–0.76) | | | |
| | 0.0009 | 0.06 | 0.03 | | | |
| High easTILs | 0.65 | 0.78 | NS | 0.54 | – | – |
| | (0.43–0.98) | (0.48–1.26) | | (0.31–0.92) | | |
| | 0.04 | 0.31 | | 0.02 | | |
| Predicted as "High risk" | – | – | – | – | 1.8 | 1.7 |
| | | | | | (1.3–2.7) | (1.2–2.6) |
| | | | | | 0.002 | 0.003 |
| | Spearman *r* coefficient *P*-value | | | | | |
| High easTILs | NS | NS | NS | 0.63[b] | – | – |
| | | | | 0.0001 | | |
| | Concordance index ("c-index") | | | | | |
| Predicted risk score (CNN output) vs. actual time-to-event data | – | – | – | – | 0.57 | 0.57 |

95% CI: 95% confidence interval. eaTILs (mm2): Density of TILs over tumor region. easTILs: Density of TILs over stroma area that mimics the international TIL working group variable as read by pathologists. eTILs%: Proportion of TILs over tumor cells. esTILs%: Proportion of TILs over stromal cells. etTILs%: Proportion of TILs over all detected cells. sTIL: Stromal TILs. TILs: Tumor-infiltrating lymphocytes.

[a] Outcomes predicted: Distant disease-free survival in patients with higher TILs scores (Bai et al. (2021)) and distant disease-free survival in patients predicted as "High risk" by ML models (by Bychkov et al. (2022)).

[b] Good correlation found when the CNN11-derived easTLs variable score was compared with the pathologist-read sTILs assessment.

images scanned with 20× and 40× magnifications. Data augmentation and color normalization were the most common preprocessing methods. Both have been widely used in computational pathology to help improve the generalizability of ML models. The first aims to increase the diversity of the training data by adding artificially generated variations of them (e.g., by rotating or mirroring the images).[69] The latter tries to reduce the effect of color variations in the images (usually a consequence of different staining or scanning processes).[69,70] Except for some similarities found in the 2 studies written by the same author,[33,34] the software platforms used for annotation/computational purposes and the information published about them were different in each study.

All the included studies, but one, used CNNs, a category of deep neural networks (DNNs). Unlike traditional computer vision approaches that require designing hand-engineered features (usually time-consuming and expensive), DNNs can learn to extract features automatically.[71] In addition this advantage, CNNs use filters (AKA convolutional kernels) that have shown to be very powerful in learning patterns from images and videos.[71] Consequently, CNNs became dominant during the last years.[72,73] Even though vision transformers have outperformed CNNs in some computer vision tasks, these started to be used more recently and were not found in any of the included studies.[20,74]

Considering that the performance of ML models when validated on external datasets usually diminishes, the results of most included studies can be regarded as encouraging. For example, all the accuracies and AUC achieved by the diagnostic or classification models were above 87% and 90%, respectively; Yang et al. obtained perfect or almost perfect precision and recall values, the prognostic models developed by Bychkov et al. and Wang et al., the hazard ratios (HR) were between 1.7 and 1.9 (with statistical significance), and all the machine TIL variables developed by Bai et al. were significantly associated with outcomes. However, as discussed below, the limited number of histological categories, classes, or types of tumors/entities in which the models were applied could restrict their usability in real-life clinical practices.

In the 2 studies where enough information was available, the risk of bias was high due to the small sample size of the validation dataset. Therefore, their predictive performance will likely be lower than that reported. Although PROBAST[42] was originally created for statistical (regression-based) prediction models, most of its items are applicable to ML-based prediction model studies.[75] As ML models need large sample sizes, insufficient sample sizes when developing and validating ML models can be considered a major design flaw.[75]

Although obtaining good results in one or more EV datasets is important evidence of generalizability,[15] it is essential to note that this doesn't guarantee that the model will perform well in all other settings.[14,76] As the amount and diversity of examples used to train ML models limit their prediction/recognition capabilities, only patterns properly represented in training datasets are expected to be adequately predicted/recognized in validation datasets.[69,76–78] According to the Food and Drug Administration (FDA) and College of American Pathologists (CAP), validation datasets must contain sufficient cases representative of those the product/algorithm will likely encounter during its intended use.[79,80] Thus, validation datasets may need to be constructed for each institution where ML models will be deployed.[14,76] Also, as they may recurrently encounter new cases with previously unseen relevant attributes, monitoring ML models' performance, and updating them (by retraining/fine-tuning their hyperparameters) recurrently could be necessary.[14,81–83]

This can become an iterative process for each institution as shown in Fig. 4.

Gathering experiences prospectively could support the future development of guidelines that define some specifics around this workflow, such as the best way to measure whether retraining of the model is needed, potential thresholds of internal/external validations, the necessity of assessing diagnostically the incorrectly predicted EV data to ensure they have representation in the training datasets and the need of pathologists' intervention to ensure that predictions are aligned with their diagnoses.

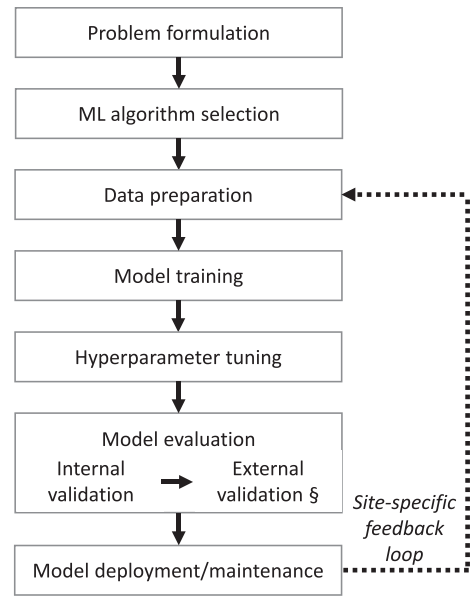

**Fig. 4.** ML models site-specific iterative development steps review.

Also, considering that ML models may become more generalizable if retrained/fine-tuned with diverse data extracted from the institutions that use them, the implications of this improvement could be discussed between these institutions and vendors when models are privately owned.[84]

*Limitations of the review process and of the evidence included in the review*

Similar to what has been described by several authors before,[46–48,63] there is an observed lack of consistency in the methods, performance metrics of the studies explored in this work. Additionally, the EVs utilized for each of the studies are heterogeneous and therefore limit the comparison across all the works and possibly inhibit a deeper understanding of how the results compare.

It is also relevant to mention that all the studies that externally validated ML models for diagnosis or classification purposes, trained, and validated the models with broad categories (or classes) of tumors/entities or with a few specific subset of them (e.g., only some histological types or subtypes). In clinical practices, pathologists always need to be able to recognize all the specific types and subtypes of tumors/entities listed in the classifications regarded as standard. The World Health Organization (WHO) classifications are the most commonly used classifications for human tumors.[85] And as an example, the class "invasive breast carcinoma" contains more than 20 specific histological types of tumors in its most recent edition (and some of them can be further subclassified in subtypes and variants).[86] Although training/testing a model to recognize all the specific types and subtypes of tumors/entities could be out of the scope of all or almost all ML models, this limitation will also undoubtedly restrict their applicability in clinical settings.

Although the search strategies were designed to be very comprehensive (See Appendix B), the studies that did not suggest, imply, or mention that their ML models underwent external validation in their Title/Abstract were not included. The same happened with those published (and indexed) in MEDLINE, EMBASE, CINAHL, IEEE, MICCAI, or SPIE conferences before January 1, 2010, or after February 28, 2022. Consequently, this systematic review did not include some ML models that might have otherwise met the inclusion criteria. For example, a study showing the results of the external validation of GALEN Breast, a commercially available algorithm used to detect invasive and in situ breast carcinomas, was excluded because it was published in December 2022.[87] Besides, numerous ML models not developed for diagnosis, classification, prognosis, or treatment outcome prediction (such as some recently reviewed by Chan et al.),[88] even if they became commercially available and were externally validated, were excluded for not meeting the inclusion criteria (e.g., had a different purpose).

Future comprehensive analyses may be facilitated with an increased availability of external validation datasets and by enhancing adherence to standardized methods and reporting protocols.

**Registration and protocol**

The review was not registered.

**Availability of data, code and other materials**

The data that support the findings of this study are available on request from the corresponding author.

**Declaration of Competing Interest**

Authors do not have any competing interests to declare. This research did not receive any specific grant from funding agencies in the public, commercial, or not-for-profit sectors.

**Acknowledgments**

The search strategy was developed in conjunction with a McMaster Health Sciences Library librarian with expertise in systematic reviews.

**Appendix A. Databases description**

*A.1. Ovid MEDLINE®*

Produced by the National Library of Medicine (NLM). It indexes information from approximately 5600 journals published world-wide, related to "biomedicine, including the allied health fields and the biological and physical sciences, humanities, and information science as they relate to medicine and health care".[89] Ovid MEDLINE® also contains PubMed-not-MEDLINE records from NLM[89] (i.e., journals/manuscripts deposited in PubMed Central and books and documents from the National Center for Biotechnology Information (NCBI) Bookshelf).[90]

*A.2. Embase (Excerpta Medica Database)*

Produced by Elsevier B.V. Indexes information from more than 8500 journals published world-wide biomedicine, and "is especially strong in its coverage of drug and pharmaceutical research, pharmacology, toxicology, and medical devices."[91]

*A.3. CINAHL*

Produced by EBSCO. Indexes the top nursing and allied health literature available from more than 3800 journals.[92]

*A.4. IEEE Xplore*

Produced by the Institute of Electrical and Electronics Engineers (IEEE), Indexes more than 6 million full-text documents (e.g., journals, conference papers, technical standards, and books) of electrical engineering, computer science, and electronics.[93]

*A.5. MICCAI*

The proceedings of the International Conference on Medical Image Computing and Computer-Assisted Intervention (MICCAI) are published in the Springer Lecture Notes in Computer Science (LNCS) series.[94] It includes oral and poster papers on "medical image computing, computer-assisted intervention, guidance systems and robotics, visualization and virtual reality, computer-aided diagnosis, bioscience and biology applications, specific imaging systems, and new imaging applications".[95]

*A.6. SPIE conferences*

Each year the Society of Photo-Optical Instrumentation Engineers (SPIE) conferences indexes 16 000 + papers and presentation recordings "reporting on photonics-driven advancements in biomedicine, astronomy, defense and security, renewable energy"[96]

**Appendix B. Search strategies**

*B.1. OVID Medline Epub Ahead of Print, In-Process & Other Non-Indexed Citations, Ovid MEDLINE(R) Daily, and Ovid MEDLINE(R) 1946 to Present*

1. "neoplasms, glandular and epithelial"/ or carcinoma/ or adenocarcinoma, mucinous/ or carcinoma, adenoid cystic/ or carcinoma, ductal/ or carcinoma, ductal, breast/ or carcinoma, lobular/ or carcinoma, mucoepidermoid/ or carcinoma, neuroendocrine/ or cystadenocarcinoma, mucinous/ or carcinoma, adenosquamous/ or carcinoma, papillary/ or carcinoma, squamous cell/ or "neoplasms, cystic, mucinous, and serous"/ or cystadenocarcinoma/ or "neoplasms, ductal, lobular, and medullary"/ or carcinoma, medullary/ or breast neoplasms/ or "hereditary breast and ovarian cancer syndrome"/ or inflammatory breast neoplasms/ or triple negative breast neoplasms/ or unilateral breast neoplasms/
2. (cancer or carcinoma$ or malignant).tw.
3. breast/ or mammary glands, human/
4. (breast or mammar$).tw.
5. 3 or 4
6. 2 and 5
7. 1 or 6
8. algorithms/ or artificial intelligence/ or machine learning/ or deep learning/ or supervised machine learning/ or support vector machine/ or unsupervised machine learning/ or decision theory/ or decision trees/ or neural networks, computer/

9. machine learning.tw. or exp algorithm/ or algorithm.tw. or Automatic Data Processing/ or Automatic Data Processing.tw. or computer aided detection.tw. or artificial intelligence.tw.
10. 8 or 9
11. pathology/ or pathology, surgical/ or histology/ or histocytochemistry/ or immunohistochemistry/
12. (histopatholog$ or patholog$ or histolog$).tw.
13. ("whole slide imag$" or WSI or (digit$ and slide$)).tw.
14. 11 or 12 or 13
15. valid*.tw.
16. diagnosis/ or clinical decision-making/ or clinical reasoning/ or diagnosis, computer-assisted/ or image interpretation, computer-assisted/ or diagnosis, differential/ or specimen handling/ or biopsy/ or biopsy, needle/ or biopsy, large-core needle/ or dissection/ or microscopy/ or photomicrography/ or diagnostic tests, routine/ or early diagnosis/ or "early detection of cancer"/ or prognosis/ or neoplasm staging/ or treatment outcome/ or disease-free survival/ or progression-free survival/ or response evaluation criteria in solid tumors/ or treatment failure/
17. (diagnos$ or classif$ or prognos$ or predict$).tw.
18. 16 or 17.

*B.2. Embase 1974 to 2022 February 28*

1. neoplasm/ or malignant neoplasm/ or solid malignant neoplasm/ or carcinoma/ or breast tumor/ or breast cancer/ or breast carcinoma/ or breast adenocarcinoma/ or metastatic breast cancer/ or ductal carcinoma/ or breast ductal carcinoma/ or lobular carcinoma/ or inflammatory breast cancer/ or metaplastic carcinoma/ or adenosquamous carcinoma/ or papillary carcinoma/ or medullary carcinoma/ or neuroendocrine carcinoma/ or adenoid cystic carcinoma/ or breast cancer molecular subtype/ or luminal A breast cancer/ or luminal B breast cancer/ or basal like breast cancer/ or triple negative breast cancer/ or estrogen receptor positive breast cancer/ or estrogen receptor negative breast cancer/ or human epidermal growth factor receptor 2 positive breast cancer/ or human epidermal growth factor receptor 2 negative breast cancer/ or "hereditary breast and ovarian cancer syndrome"/
2. (cancer or carcinoma$ or malignant).tw.
3. breast/ or mammary gland/
4. (breast or mammar$).tw.
5. 3 or 4
6. 2 and 5
7. 1 or 6
8. machine learning/ or artificial neural network/ or computer vision/ or automated pattern recognition/ or back propagation/ or bayesian learning/ or classification algorithm/ or classifier/ or clustering algorithm/ or computer heuristics/ or convolution algorithm/ or cross validation/ or data mining/ or deconvolution algorithm/ or detection algorithm/ or Dijkstra's algorithm/ or dimensionality reduction/ or dynamic time warping/ or empirical mode decomposition/ or feature detection/ or feature extraction/ or feature extraction algorithm/ or "feature learning (machine learning)"/ or feature ranking/ or feature selection/ or feature selection algorithm/ or fuzzy system/ or generalized method of moments/ or greedy algorithm/ or hidden markov model/ or imaging algorithm/ or iterative closest point/ or k nearest neighbor/ or kernel method/ or knowledge discovery/ or learning algorithm/ or Levenberg Marquardt algorithm/ or Markov jump/ or markov state model/ or maximum entropy model/ or maximum likelihood method/ or Metropolis Hastings algorithm/ or model predictive control/ or multicriteria decision analysis/ or multifactor dimensionality reduction/ or Needleman Wunsch algorithm/ or network learning/ or online analytical processing/ or perceptron/ or radial basis function/ or radial basis function/ or random forest/ or recursive feature elimination/ or recursive partitioning/ or relevance vector machine/ or risk algorithm/ or rough set/ or successive projections algorithm/ or semi supervised machine learning/ or superposition algorithm/ or supervised machine learning/ or support vector machine/ or unsupervised machine learning/
9. machine learning.mp. or machine learning/ or algorithm.mp. or algorithm/ or information processing.mp. or information processing/ or artificial intelligence.mp. or artificial intelligence/
10. 8 or 9
11. pathology/ or general pathology/ or histopathology/ or molecular pathology/ or pathological anatomy/ or neuropathology/ or histology/ or histometry/ or histophotometry/ or brain histology/ or liver histology/ or immunohistochemistry.mp. [mp=title, abstract, heading word, drug trade name, original title, device manufacturer, drug manufacturer, device trade name, keyword heading word, floating subheading word, candidate term word]
12. (histopatholog$ or patholog$ or histolog$).tw.
13. ("whole slide imag$" or WSI or (digit$ and slide$)).tw.
14. 11 or 12 or 13
15. valid$.tw.
16. diagnosis/ or early diagnosis/ or tumor diagnosis/ or cancer diagnosis/ or early cancer diagnosis/ or differential diagnosis/ or diagnosis related group/ or tumor biopsy/ or diagnostic test/ or cancer test/ or laboratory diagnosis/ or molecular diagnosis/ or computer assisted diagnosis/ or quantitative diagnosis/ or diagnostic accuracy/ or diagnostic test accuracy study/ or diagnostic error/ or missed diagnosis/ or diagnostic reasoning/ or cancer grading/ or cancer staging/ or prognosis/ or cancer prognosis/ or disease course/ or survival/ or survival prediction/ or cancer survival/ or cancer specific survival/ or survival rate/ or mean survival time/ or median survival time/ or overall survival/ or survival analysis/ or long term survival/ or cancer free survival/ or disease free survival/ or disease specific survival/ or disease free interval/ or event free survival/ or progression free survival/ or recurrence free survival/ or local progression free survival/ or distant progression free survival/ or local recurrence free survival/ or distant recurrence free survival/ or local disease free survival/ or distant disease free survival/ or metastasis free survival/ or distant metastasis free survival/ or failure free survival/ or local failure free survival/ or regional failure free survival/ or distant failure free survival/ or treatment free survival/ or post treatment survival/ or cancer recurrence/ or cancer regression/ or tumor recurrence/ or tumor regression/ or recurrent disease/ or relapse/ or remission/ or prediction/ or computer prediction/ or forecasting/ or predictive validity/ or predictive value/ or outcome assessment/ or response evaluation criteria in solid tumors/ or treatment response/ or treatment response time/ or treatment failure/ or treatment outcome/ or disease worsening with drug treatment/ or outcomes research/

17. (diagnos$ or classif$ or prognos$ or predict$).tw.
18. 16 or 17
19. 7 and 10 and 14 and 15 and 18.

*B.3. CINAHL. Interface: EBSCOhost Research Databases*

( ((MH "Neoplasms, Glandular and Epithelial") OR (MH "Carcinoma") OR (MH "Carcinoma, Adenoid Cystic") OR (MH "Carcinoma, Ductal") OR (MH "Carcinoma, Ductal, Breast") OR (MH "Carcinoma, Lobular") OR (MH "Carcinoma, Neuroendocrine") OR (MH "Carcinoma, Papillary") OR (MH "Carcinoma, Squamous Cell") OR (MH "Neoplasms, Cystic, Mucinous, and Serous") OR (MH "Neoplasms, Ductal, Lobular, and Medullary") OR (MH "Breast Neoplasms") OR (MH "Hereditary Breast and Ovarian Cancer Syndrome") OR (((MH "Breast") OR (TI breast OR AB breast) OR (TI mammar* OR AB mammar*)) AND ((TI cancer OR AB cancer) OR (TI carcinoma* OR AB carcinoma*) OR (TI malignant OR AB malignant)))) ) AND ( ((MH "Algorithms") OR (MH "Artificial Intelligence") OR (MH "Machine Learning") OR (MH "Deep Learning") OR (MH "Support Vector Machine") OR (MH "Decision Trees") OR (MH "Neural Networks (Computer)") OR (TI "machine learning" OR AB "machine learning") OR (TI algorithm OR AB algorithm) OR (TI "Automatic Data Processing" OR AB "Automatic Data Processing") OR (TI "computer aided detection" OR AB "computer aided detection") OR (TI "artificial intelligence" OR AB "artificial intelligence")) ) AND ( ((MH "Pathology") OR (MH "Histology") OR (MH "Histocytochemistry") OR (MH "Immunohistochemistry") OR (TI histopathology* OR AB histopathology*) OR (TI patholog* OR AB patholog*) OR (TI histolog* OR AB histolog*) or ("whole slide imag*" OR WSI OR (digit* AND slide))) ) AND ( (TI valid* OR AB valid*) ) OR ( ((MH "Diagnosis") OR (MH "Decision Making, Clinical") OR (MH "Clinical Reasoning") OR (MH "Diagnosis, Computer Assisted") OR (MH "Image Interpretation, Computer Assisted") OR (MH "Diagnosis, Differential") OR (MH "Specimen Handling") OR (MH "Biopsy") OR (MH "Biopsy, Needle") OR (MH "Dissection") OR (MH "Microscopy") OR (MH "Diagnostic Tests, Routine") OR (MH "Early Diagnosis") OR (MH "Early Detection of Cancer") OR (MH "Prognosis") OR (MH "Neoplasm Staging") OR (MH "Treatment Outcomes") OR (MH "Treatment Failure") OR (TI diagnos* OR AB diagnos*) OR (TI diagnos* OR AB diagnos*) OR (TI classif* OR AB classif*) OR (TI prognos* OR AB prognos*) or (TI predict* OR AB predict*)) ).

*B.4. IEEE via IEEE Xplore*

((("Mesh_Terms":"neoplasms, glandular and epithelial" OR "Mesh_Terms":carcinoma OR "Mesh_Terms":adenocarcinoma, mucinous OR "Mesh_Terms":carcinoma, adenoid cystic OR "Mesh_Terms":carcinoma, ductal OR "Mesh_Terms":carcinoma, ductal, breast OR "Mesh_Terms":carcinoma, lobular OR "Mesh_Terms":carcinoma, mucoepidermoid OR "Mesh_Terms":carcinoma, neuroendocrine OR "Mesh_Terms":cystadenocarcinoma, mucinous OR "Mesh_Terms":carcinoma, adenosquamous OR "Mesh_Terms":carcinoma, papillary OR "Mesh_Terms":carcinoma, squamous cell OR "Mesh_Terms":"neoplasms, cystic, mucinous, and serous" OR "Mesh_Terms":cystadenocarcinoma OR "Mesh_Terms":"neoplasms, ductal, lobular, and medullary" OR "Mesh_Terms":carcinoma, medullary OR "Mesh_Terms":breast neoplasms OR "Mesh_Terms":"hereditary breast and ovarian cancer syndrome" OR "Mesh_Terms":inflammatory breast neoplasms OR "Mesh_Terms":triple negative breast neoplasms OR "Mesh_Terms":unilateral breast neoplasms) OR ((("Index Terms":cancer OR "Index Terms":carcinoma OR "Index Terms":malignant) OR ("Abstract":cancer OR "Abstract":carcinoma OR "Abstract":malignant)) AND (("Mesh_Terms":breast OR "Mesh_Terms":mammary glands, human) OR ("Index Terms":breast OR "Index Terms":mammary OR "Index Terms":mammarian) OR ("Abstract":breast OR "Abstract":mammary OR "Abstract":mammarian)))) AND (("Mesh_Terms":algorithms OR "Mesh_Terms":artificial intelligence OR "Mesh_Terms":machine learning OR "Mesh_Terms":deep learning OR "Mesh_Terms":supervised machine learning OR "Mesh_Terms":support vector machine OR "Mesh_Terms":unsupervised machine learning OR "Mesh_Terms":decision theory OR "Mesh_Terms":decision trees OR "Mesh_Terms":neural networks, computer) OR ("IEEE Terms":Machine learning algorithms OR "IEEE Terms":Artificial intelligence OR "IEEE Terms":Machine learning OR "IEEE Terms":Deep learning OR "IEEE Terms":Supervised learning OR "IEEE Terms":Unsupervised learning OR "IEEE Terms":Semisupervised learning OR "IEEE Terms":Support vector machines OR "IEEE Terms":Artificial neural networks) OR ("Index Terms":machine learning OR "Index Terms":artificial intelligence) OR ("Abstract":machine learning)) AND (("Mesh_Terms":pathology OR "Mesh_Terms":pathology, surgical OR "Mesh_Terms":histology OR "Mesh_Terms":histocytochemistry OR "Mesh_Terms":immunohistochemistry) OR ("Index Terms":histopathology OR "Index Terms":histopathological OR "Index Terms":pathology OR "Index Terms":pathological OR "Index Terms":histology OR "Index Terms":histological) OR ("Abstract":histopathology OR "Abstract":histopathological OR "Abstract":pathology OR "Abstract":pathological OR "Abstract":histology OR "Abstract":histological) OR ("Index Terms":"whole slide images" OR "Index Terms":"whole slide imaging" OR "Index Terms":WSIs) OR ("Abstract":"whole slide images" OR "Abstract":"whole slide imaging" OR "Abstract":WSIs) OR (("Index Terms":digital OR "Index Terms":digitizing OR "Index Terms":digitized) AND ("Index Terms":slides)) OR (("Abstract":digital OR "Abstract":digitizing OR "Abstract":digitized) AND ("Abstract":slides))) AND (("Index Terms":valid*) OR ("Abstract":valid*)).

*B.5. MICCAI via Springer link*

(Breast OR mammary) AND (Cancer OR carcinoma OR tumor OR tumors OR tumour OR tumours OR neoplasm OR neoplasms OR malignant) AND ("artificial intelligence" OR "machine Learning" OR "Deep learning" OR "neural network" OR "computer vision" OR Algorithm OR Algorithms) AND (pathology OR histology OR histopathology OR histological OR histopathological OR "whole slide images" OR "whole slide imaging" OR WSI OR WSIs OR "digital slides") AND (valid*). Filters: within miccai & Conference Paper & 2010–2022.

*B.6. SPIE conferences*

ABSTRACT:(Breast OR mammary) AND ABSTRACT:(Cancer OR carcinoma OR tumor OR tumors OR tumour OR tumours OR neoplasm OR neoplasms OR malignant) AND ABSTRACT:("artificial intelligence" OR "machine Learning" OR "Deep learning" OR "neural network" OR "computer vision" OR Algorithm OR Algorithms) AND ABSTRACT:(pathology OR histology OR histopathology OR histological OR histopathological OR "whole slide images" OR "whole slide imaging" OR WSI OR WSIs OR "digital slides") AND ABSTRACT:(validated OR validation OR valid OR validating OR validate). Filters: Proceedings & 2010–2022.

## Appendix C. Examples of records that passed and did not pass the Title/Abstract screening

**Improved breast cancer histological grading using deep learning**

Background: The Nottingham histological grade (NHG) is a well-established prognostic factor for breast cancer that is broadly used in clinical decision making. However, ~50% of patients are classified as grade 2, an intermediate risk group with low clinical value. To improve risk stratification of NHG 2 breast cancer patients, we developed and validated a novel histological grade model (DeepGrade) based on digital whole-slide histopathology images (WSIs) and deep learning. Patients and Methods: In this observational retrospective study, routine WSIs stained with haematoxylin and eosin from 1567 patients were utilised for model optimisation and validation. Model generalisability was further evaluated in an external test set with 1262 patients. NHG 2 cases were stratified into two groups, DG2-high and DG2-low, and the prognostic value was assessed. The main outcome was recurrence-free survival. Result(s): DeepGrade provides independent prognostic information for stratification of NHG 2 cases in the internal test set, where DG2-high showed an increased risk for recurrence (hazard ratio [HR] 2.94, 95% confidence interval [CI] 1.24-6.97, P = 0.015) compared with the DG2-low group after adjusting for established risk factors (independent test data). DG2-low also shared phenotypic similarities with NHG 1, and DG2-high with NHG 3, suggesting that the model identifies morphological patterns in NHG 2 that are associated with more aggressive tumours. The prognostic value of DeepGrade was further assessed in the external test set, confirming an increased risk for recurrence in DG2-high (HR 1.91, 95% CI 1.11-3.29, P = 0.019). Conclusion(s): The proposed model-based stratification of patients with NHG 2 tumours is prognostic and adds clinically relevant information over routine histological grading. The methodology offers a cost-effective alternative to molecular profiling to extract information relevant for clinical decisions. Copyright © 2021 The Authors

**Authors:** Wang Y.; Acs B.; Robertson S.; Liu B.; Solorzano L.; Wahlby C.; Hartman J.; Rantalainen M.;

**Journal:** Annals of Oncology - Volume 33, Issue 1, pp. 89-98 - published 2022-01-01

**Publication Types:** Journal Article

**Fig. 5.** Record that passed the Title/Abstract screening and the full-text screening.

**Outcome and Biomarker Supervised Deep Learning for Survival Prediction in Two Multicenter Breast Cancer Series.**

Background: Prediction of clinical outcomes for individual cancer patients is an important step in the disease diagnosis and subsequently guides the treatment and patient counseling. In this work, we develop and evaluate a joint outcome and biomarker supervised (estrogen receptor expression and ERBB2 expression and gene amplification) multitask deep learning model for prediction of outcome in breast cancer patients in two nation-wide multicenter studies in Finland (the FinProg and FinHer studies). Our approach combines deep learning with expert knowledge to provide more accurate, robust, and integrated prediction of breast cancer outcomes., Materials and Methods: Using deep learning, we trained convolutional neural networks (CNNs) with digitized tissue microarray (TMA) samples of primary hematoxylin-eosin-stained breast cancer specimens from 693 patients in the FinProg series as input and breast cancer-specific survival as the endpoint. The trained algorithms were tested on 354 TMA patient samples in the same series. An independent set of whole-slide (WS) tumor samples from 674 patients in another multicenter study (FinHer) was used to validate and verify the generalization of the outcome prediction based on CNN models by Cox survival regression and concordance index (c-index). Visual cancer tissue characterization, i.e., number of mitoses, tubules, nuclear pleomorphism, tumor-infiltrating lymphocytes, and necrosis was performed on TMA samples in the FinProg test set by a pathologist and combined with deep learning-based outcome prediction in a multitask algorithm., Results: The multitask algorithm achieved a hazard ratio (HR) of 2.0 (95% confidence interval [CI] 1.30-3.00), P < 0.001, c-index of 0.59 on the 354 test set of FinProg patients, and an HR of 1.7 (95% CI 1.2-2.6), P = 0.003, c-index 0.57 on the WS tumor samples from 674 patients in the independent FinHer series. The multitask CNN remained a statistically independent predictor of survival in both test sets when adjusted for histological grade, tumor size, and axillary lymph node status in a multivariate Cox analyses. An improved accuracy (c-index 0.66) was achieved when deep learning was combined with the tissue characteristics assessed visually by a pathologist., Conclusions: A multitask deep learning algorithm supervised by both patient outcome and biomarker status learned features in basic tissue morphology predictive of survival in a nationwide, multicenter series of patients with breast cancer. The algorithms generalized to another independent multicenter patient series and whole-slide breast cancer samples and provide prognostic information complementary to that of a comprehensive series of established prognostic factors. Copyright: © 2022 Journal of Pathology Informatics.

**Authors:** Bychkov, Dmitrii; Joensuu, Heikki; Nordling, Stig; Tiulpin, Aleksei; Kucukel, Hakan; Lundin, Mikael; Sihto, Harri; Isola, Jorma; Lehtimaki, Tiina; Kellokumpu-Lehtinen, Pirkko-Liisa; von Smitten, Karl; Lundin, Johan; Linder, Nina;

**Journal:** Journal of pathology informatics - Volume 13, Issue 101528849, pp. 9 - published 2022-01-01

**Publication Types:** Journal Article

**Fig. 6.** Record that passed the Title/Abstract screening and the full-text screening.

**Automated detection and grading of Invasive Ductal Carcinoma breast cancer using ensemble of deep learning models.**

Invasive ductal carcinoma (IDC) breast cancer is a significant health concern for women all around the world and early detection of the disease may increase the survival rate in patients. Therefore, Computer-Aided Diagnosis (CAD) based systems can assist pathologists to detect the disease early. In this study, we present an ensemble model to detect IDC using DenseNet-121 and DenseNet-169 followed by test time augmentation (TTA). The model achieved a balanced accuracy of 92.70% and an F1-score of 95.70% outperforming the current state-of-the-art. Comparative analysis against various pre-trained deep learning models and preprocessing methods have been carried out. Qualitative analysis has also been conducted on the test dataset. After the detection of IDC breast cancer, it is important to grade it for further treatment. In our study, we also propose an ensemble model for the grading of IDC using the pre-trained DenseNet-121, DenseNet-201, ResNet-101v2, and ResNet-50 architectures. The model is inferred from two validation cohorts. For the patch-level classification, the model yielded an overall accuracy of 69.31%, 75.07%, 61.85%, and 60.50% on one validation cohort and 62.44%, 79.14%, 76.62%, and 71.05% on the second validation cohort for 4x, 10x, 20x, and 40x magnified images respectively. The same architecture is further validated using a different IDC dataset where it achieved an overall accuracy of 90.07%. The performance of the models on the detection and grading of IDC shows that they can be useful to help pathologists detect and grade the disease. Copyright © 2021 Elsevier Ltd. All rights reserved.

**Authors:** Barsha, Nusrat Ameen; Rahman, Aimon; Mahdy, M R C;

**Journal:** Computers in biology and medicine - Volume 139, Issue 0, pp. 104931 - published 2021-01-01

**Publication Types:** Journal Article

**Fig. 7.** Record that passed the Title/Abstract screening but did not pass the full-text screening (because EV was not performed).

> **HWDCNN: Multi-class recognition in breast histopathology with Haar wavelet decomposed image based convolution neural network**
>
> Among the predominant cancers, breast cancer is one of the main causes of cancer deaths impacting women worldwide. However, breast cancer classification is challenging due to numerous morphological and textural variations that appeared in intra-class images. Also, the direct processing of high resolution histological images is uneconomical in terms of GPU memory. In the present study, we have proposed a new approach for breast histopathological image classification that uses a deep convolution neural network (CNN) with wavelet decomposed images. The original microscopic image patches of 2048 x 1536 x 3 pixels are decomposed into 512 x 384 x 3 using 2-level Haar wavelet and subsequently used in proposed CNN model. The image decomposition step considerably reduces convolution time in deep CNNs and computational resources, without any performance downgrade. The CNN model extracts the deep features from Haar wavelet decomposed images and incorporates multi-scale discriminant features for precise prognostication of class labels. This paper also solves the demand for massive histopathology dataset by means of transfer learning and data augmentation techniques. The efficacy of proposed approach is corroborated on two publicly available breast histology datasets-(a) one provided as a part of international conference on image analysis and recognition (ICIAR 2018) grand challenge and (b) on BreakHis data. On the ICIAR 2018 validation data, our model showed an accuracy of 98.2% for both 4-class and 2-class recognition. Further, on hidden test data of the ICIAR 2018, we achieved an accuracy of 91%, outperforming existing state-of-the-art results significantly. Besides, on BreakHis dataset, the model achieved competing performance with 96.85% multi-class accuracy.Copyright © 2019
>
> **Authors:** Kausar T.; Wang M.; Idrees M.; Lu Y.;
>
> **Journal:** Biocybernetics and Biomedical Engineering - Volume 39, Issue 4, pp. 967-982 - published 2019-01-01
>
> **Publication Types:** Journal Article

**Fig. 8.** Record that passed the Title/Abstract screening but did not pass the full-text screening (because EV was not performed).

> **Morphological subtyping of breast cancer using machine learning**
>
> Background: Breast pathology includes a spectrum of benign, atypical, preinvasive and invasive lesions. Breast biopsy diagnoses are challenging and without expert review have a relatively high discordance rate in interpretation. Due to routine screening, breast specimens generate high volumes in pathology, and lesion detection and distinct classification are important. Machine learning (ML) has been shown to improve efficiency and accuracy of cancer detection in a variety of tissue types. Purpose(s): We hypothesized that ML can be trained to detect and subclassify clinically meaningful lesions in breast pathology with the potential to provide increased efficiency and accuracy to the histologic examination of slides of breast tissue. Method(s): De-identified glass slides were scanned on Leica AT2 whole slide scanners (0.5 mum/pixel). Diagnostic classes were available from the pathology report. Using a machine learning architecture of multiple instance learning, an SE-ResNet50 Convolutional Neural Network (CNN) was trained. Each input to the network consisted of whole slide images (WSIs). Classification performances for invasive carcinoma, ductal carcinoma in situ (DCIS), lobular carcinoma in situ (LCIS), atypical ductal hyperplasia (ADH), and atypical lobular hyperplasia (ALH) were determined. Result(s): A dataset of 9,751 anatomical specimens (biopsy, 6,289; excision, 3,462) comprising 40,637 slides were used to train the CNN. The system was validated on WSIs generated from 3,742 breast specimens (biopsy, 2,250; excision, 1,492) comprising 13,601 digital slides that were not included in the training of the CNN model. Results shows area under the receiver operating characteristic curve (AUC) classification of invasive carcinoma, DCIS, LCIS, ALH, and ADH as 0.98, 0.98, 0.97, 0.95, and 0.92, respectively. Conclusion(s): The trained CNN had very high performance in classifying breast lesions including invasive carcinoma, DCIS, LCIS, ADH, and ALH. Future work includes distinguishing between invasive lobular and invasive ductal carcinoma, as well as generalisability studies with slides prepared at other institutions..
>
> **Authors:** Hanna M.; Lee M.; Bozkurt A.; Hamilton P.P.; Godrich R.; Casson A.; Raciti P.; Sue J.; Viret J.; Lee D.; Grady L.; Rothrock B.; Dogdas B.; Fuchs T.; Reis-Filho J.; Kanan C.;
>
> **Journal:** Journal of Pathology - Volume 255, Issue 0, pp. S35 - published 2021-01-01
>
> **Publication Types:** Journal Article

**Fig. 9.** Record that did not pass the Title/Abstract screening (because EV was not performed).

> **Cancer Grade Model: a multi-gene machine learning-based risk classification for improving prognosis in breast cancer.**
>
> BACKGROUND: Prognostic stratification of breast cancers remains a challenge to improve clinical decision making. We employ machine learning on breast cancer transcriptomics from multiple studies to link the expression of specific genes to histological grade and classify tumours into a more or less aggressive prognostic type., MATERIALS AND METHODS: Microarray data of 5031 untreated breast tumours spanning 33 published datasets and corresponding clinical data were integrated. A machine learning model based on gradient boosted trees was trained on histological grade-1 and grade-3 samples. The resulting predictive model (Cancer Grade Model, CGM) was applied on samples of grade-2 and unknown-grade (3029) for prognostic risk classification., RESULTS: A 70-gene signature for assessing clinical risk was identified and was shown to be 90% accurate when tested on known histological-grade samples. The predictive framework was validated through survival analysis and showed robust prognostic performance. CGM was cross-referenced with existing genomic tests and demonstrated the competitive predictive power of tumour risk., CONCLUSIONS: CGM is able to classify tumours into better-defined prognostic categories without employing information on tumour size, stage, or subgroups. The model offers means to improve prognosis and support the clinical decision and precision treatments, thereby potentially contributing to preventing underdiagnosis of high-risk tumours and minimising over-treatment of low-risk disease. Copyright © 2021. The Author(s).
>
> **Authors:** Amiri Souri, E; Chenoweth, A; Cheung, A; Karagiannis, S N; Tsoka, S;
>
> **Journal:** British journal of cancer - Volume 125, Issue 5, pp. 748-758 - published 2021-01-01
>
> **Publication Types:** Journal Article

**Fig. 10.** Record that did not pass the Title/Abstract screening (because the model was only trained with microarray and clinical data).

> **Ultrasound-based deep learning radiomics in the assessment of pathological complete response to neoadjuvant chemotherapy in locally advanced breast cancer.**
>
> The aim of the study was to develop and validate a deep learning radiomic nomogram (DLRN) for preoperatively assessing breast cancer pathological complete response (pCR) after neoadjuvant chemotherapy (NAC) based on the pre- and post-treatment ultrasound. Patients with locally advanced breast cancer (LABC) proved by biopsy who proceeded to undergo preoperative NAC were enrolled from hospital #1 (training cohort, 356 cases) and hospital #2 (independent external validation cohort, 236 cases). Deep learning and handcrafted radiomic features reflecting the phenotypes of the pre-treatment (radiomic signature [RS] 1) and post-treatment tumour (RS2) were extracted. The minimum redundancy maximum relevance algorithm and the least absolute shrinkage and selection operator regression were used for feature selection and RS construction. A DLRN was then developed based on the RSs and independent clinicopathological risk factors. The performance of the model was assessed with regard to calibration, discrimination and clinical usefulness. The DLRN predicted the pCR status with accuracy, yielded an area under the receiver operator characteristic curve of 0.94 (95% confidence interval, 0.91–0.97) in the validation cohort, with good calibration. The DLRN outperformed the clinical model and single RS within both cohorts (P < 0.05, as per the DeLong test) and performed better than two experts' prediction of pCR (both P < 0.01 for comparison of total accuracy). Besides, prediction within the hormone receptor–positive/human epidermal growth factor receptor 2 (HER2)–negative, HER2+ and triple-negative subgroups also achieved good discrimination performance, with an AUC of 0.90, 0.95 and 0.93, respectively, in the external validation cohort. Decision curve analysis confirmed that the model was clinically useful. A deep learning–based radiomic nomogram had good predictive value for accurate pCR in LABC, which could provide valuable information for individual treatment. • A novel preoperative pathological complete response prediction model was developed. • The model is based on pre- and post-neoadjuvant chemotherapy ultrasound images. • It yielded an area under the receiver operator characteristic curve AUC of 0.94 in the independent external validation cohort. • It outperformed two experts who evaluated the pathological complete response status. • It may facilitate tailoring the optimum extent of breast and axillary surgery.
>
> **Authors:** Jiang, Meng; Li, Chang-Li; Luo, Xiao-Mao; Chuan, Zhi-Rui; Lv, Wen-Zhi; Li, Xu; Cui, Xin-Wu; Dietrich, Christoph F.;
>
> **Journal:** European Journal of Cancer - Volume 147, Issue 0, pp. 95-105 - published 2021-01-01
>
> **Publication Types:** Journal Article

**Fig. 11.** Record that did not pass the Title/Abstract screening (because the model was only trained with ultrasound images).

> **Multi-Scale Deep Neural Network for Mitosis Detection in Histological Images**
>
> Mitotic figure detection in breast cancer images plays an important role to measure aggressiveness of the cancer tumor. Currently, in clinic environment the pathologist visualized the multiple high power fields (HPFs) on a glass slide under super microscope which is an extremely tedious and time consuming process. Development of the automatic mitotic detection methods is need of time, however it also bears, scale invariance, deficiency of data, improper image staining and sample class unbalanced dilemma. These limitations are however; prohibit the automatic histopathology image analysis to be applied in clinical practice. In this paper, an automatic domain agnostic deep multi-scale fused fully convolutional neural network (MFF-CNN) is presented to detect mitoses in Hematoxylin and eosin (H&E) images. The intended model fuses the multi-level and multi-scale features and context information for accurate mitotic count and in training phase multi-step fine-tuning strategy is used to reduce the over-fitting. Moreover, the training image samples efficiently built by stain normalized the poorly stained (H&E) images and by applying an automatic sample selection strategy. Preliminarily validation on the public MITOS-ATYPIA-14 challenge dataset, demonstrate the efficiency of proposed work. The proposed method achieves better performance in term of detection accuracy with an acceptable detection speed compared to other state-of-the-art designs.
>
> **Authors:** T. Kausar; M. Wang; B. Wu; M. Idrees; B. Kanwal;
>
> **Journal:** 2018 International Conference on Intelligent Informatics and Biomedical Sciences (ICIIBMS) - Volume 3, Issue 0, pp. 47-51 - published 2018-01-01
>
> **Publication Types:** CONF

**Fig. 12.** Record that did not pass the Title/Abstract screening (because the model was developed to detect mitoses).

> **Artificial Intelligence Algorithms to Assess Hormonal Status From Tissue Microarrays in Patients With Breast Cancer.**
>
> Importance: Immunohistochemistry (IHC) is the most widely used assay for identification of molecular biomarkers. However, IHC is time consuming and costly, depends on tissue-handling protocols, and relies on pathologists' subjective interpretation. Image analysis by machine learning is gaining ground for various applications in pathology but has not been proposed to replace chemical-based assays for molecular detection., Objective: To assess the prediction feasibility of molecular expression of biomarkers in cancer tissues, relying only on tissue architecture as seen in digitized hematoxylin-eosin (H&E)-stained specimens., Design, Setting, and Participants: This single-institution retrospective diagnostic study assessed the breast cancer tissue microarrays library of patients from Vancouver General Hospital, British Columbia, Canada. The study and analysis were conducted from July 1, 2015, through July 1, 2018. A machine learning method, termed morphological-based molecular profiling (MBMP), was developed. Logistic regression was used to explore correlations between histomorphology and biomarker expression, and a deep convolutional neural network was used to predict the biomarker expression in examined tissues., Main Outcomes and Measures: Positive predictive value (PPV), negative predictive value (NPV), and area under the receiver operating characteristics curve measures of MBMP for assessment of molecular biomarkers., Results: The database consisted of 20600 digitized, publicly available H&E-stained sections of 5356 patients with breast cancer from 2 cohorts. The median age at diagnosis was 61 years for cohort 1 (412 patients) and 62 years for cohort 2 (4944 patients), and the median follow-up was 12.0 years and 12.4 years, respectively. Tissue histomorphology was significantly correlated with the molecular expression of all 19 biomarkers assayed, including estrogen receptor (ER), progesterone receptor (PR), and ERBB2 (formerly HER2). Expression of ER was predicted for 105 of 207 validation patients in cohort 1 (50.7%) and 1059 of 2046 validation patients in cohort 2 (51.8%), with PPVs of 97% and 98%, respectively, NPVs of 68% and 76%, respectively, and accuracy of 91% and 92%, respectively, which were noninferior to traditional IHC (PPV, 91%-98%; NPV, 51%-78%; and accuracy, 81%-90%). Diagnostic accuracy improved given more data. Morphological analysis of patients with ER-negative/PR-positive status by IHC revealed resemblance to patients with ER-positive status (Bhattacharyya distance, 0.03) and not those with ER-negative/PR-negative status (Bhattacharyya distance, 0.25). This suggests a false-negative IHC finding and warrants antihormonal therapy for these patients., Conclusions and Relevance: For at least half of the patients in this study, MBMP appeared to predict biomarker expression with noninferiority to IHC. Results suggest that prediction accuracy is likely to improve as data used for training expand. Morphological-based molecular profiling could be used as a general approach for mass-scale molecular profiling based on digitized H&E-stained images, allowing quick, accurate, and inexpensive methods for simultaneous profiling of multiple biomarkers in cancer tissues.
>
> **Authors:** Shamai, Gil; Binenbaum, Yoav; Slossberg, Ron; Duek, Irit; Gil, Ziv; Kimmel, Ron;
>
> **Journal:** JAMA network open - Volume 2, Issue 7, pp. e197700 - published 2019-01-01
>
> **Publication Types:** Journal Article

**Fig. 13.** Record that did not pass the Title/Abstract screening (because the model was developed to predict the expression of biomarkers).

**Appendix D. Algorithms**

| Authors | Algorithms |
|---|---|
| Cano et al. | IDC-CNN 9L-16n: 9 layers: 3 convolutional-pooling layers of 16 neurons per layer, 3 convolutional-pooling layers of 32 neurons per layer, 1 fully connected layer of 16 neurons, 1 fully connected layer of 32 neurons, and 1 sigmoid classification layer |
| Cruz-Roa et al. | A 3-layer CNN. One convolutional and pooling layer and a fully connected layer (each with 256 neurons). The third one was a classification layer, with 2 units as outputs (one for each class; invasive and non-invasive) |
| Cruz-Roa et al. | CS256-FC256: One convolutional and subsampling/pooling layer (with 256 neurons), a fully connected layer (with 256 neurons), and a final classification layer (Softmax classifier) |
| Colon-Cartagena et al. | - ResNet-50 model <br> - A coding-free custom image classifier and visual recognition model (created with the IBM Watson Visual Recognition platform) |
| Mi et al | Inception V3 was selected as the patch-level classifier. A full connection layer (with 1024 neurons) and a Softmax layer were added after the basic model |
| Radiya-Dixit et al | Best performance model: Combined model with Active Feature Extraction - CAFE; consisting of two logistic regression algorithms |
| Yang et al | Nine fine-tuned deep CNN were obtained using ResNet-101, ResNet-152, and DenseNet-161 as backbone networks. The combination of the 3 of them achieved the best performance |
| Bai et al | CNN11 TIL algorithm (Neural network with 8 hidden layers) |
| Bychkov et al. | Deep learning model around a ResNet CNN backbone (with a dropout layer introduced before the fully connected blocks) |
| Wang et al. | InceptionV3 (with weights initialized from a model pretrained by ImageNet) |

**Appendix E. Results of internal validation of ML models**

**Table 5**
Results of internal validation of diagnostic models.

| Study | Cano et al. (2018) | Cruz Roa et al. (2017) | | Cruz-Roa et al. (2018) |
|---|---|---|---|---|
| Dataset | | ConvNet$_{HUP}$ | ConvNet$_{UHCMC/CWRU}$ | |
| Accuracy (%) | 90.07 | – | – | – |
| Dice coefficient(%) | – | 67.71 | 65.96 | 67 |
| Positive-predictive value (%) | – | 64.64 | 63.70 | – |
| Negative-predictive value (%) | – | 97.09 | 96.63 | – |
| True-positive rate (%) | – | – | – | – |
| True-negative rate (%) | – | – | – | – |
| False-positive rate (%) | – | – | – | – |
| False-negative rate (%) | – | – | – | – |
| Heatmaps | – | – | – | – |

ConvNet$_{HUP}$: Convolutional Neural Network trained with cases from the Hospital of the University of Pennsylvania. ConvNet$_{UHCMC/CWRU}$: Convolutional Neural Network trained with cases from the University Hospitals Case Medical Center and Case Western Reserve University.

**Table 6**
Results of internal validation of classification models.

| Study | Colon-Cartagena et al. (2020) | Mi et al. (2021) | | | | Radiya-Dixit et al. (2017) | Yang et al. (2019) | | | | |
|---|---|---|---|---|---|---|---|---|---|---|---|
| Dataset/category | | Normal | Benign | In situ | Invasive | | Normal | Benign | In situ | Invasive | All |
| Accuracy (%) | 90 | 86.67 | | | | – | 95.98 | 94.24 | 96.01 | 97.26 | 91.75 |
| Area under the curve (%) | – | 97.56 | 96.62 | 98.37 | 95.15 | 92.1 | 98.78 | 96.40 | 98.38 | 98.50 | – |
| Precision (%) | – | 80.77 | 80.95 | 94.41 | 92.88 | – | 89.27 | 97.65 | 91.36 | 92.57 | – |
| Recall (%) | – | 92.75 | 83.12 | 87.78 | 82.75 | – | 97 | 79 | 94 | 97 | – |
| Heatmaps | – | – | – | – | – | – | – | – | – | – | – |

**Table 7**
Results of internal validation of prognostic models.

| Study | Bai et al. (2021) | Bychkov et al. (2022) | |
|---|---|---|---|
| Dataset/Model | WTS Yale | "Solo" model | Multitask model |
| | Hazard ratios[a] <br> (95% CI) <br> *P*-value | | |
| High eTILs% | 0.35 <br> (0.20–0.61) <br> 0.0002 | – | – |
| High etTILS% | 0.35 <br> (0.19–0.63) <br> 0.0005 | – | – |
| High esTILs | 0.35 <br> (0.18–0.65) <br> 0.001 | – | – |
| High eaTILs (mm2) | 0.35 | – | – |

Table 7 (*continued*)

| Study | Bai et al. (2021) | Bychkov et al. (2022) | |
| --- | --- | --- | --- |
| Dataset/Model | WTS Yale | "Solo" model | Multitask model |
|  | (0.20–0.63) |  |  |
|  | 0.0005 |  |  |
|  | 0.30 |  |  |
| High easTILs | (0.16-0.54) | – | – |
|  | <0.0001 |  |  |
|  |  | 1.7 | 2 |
| Predicted as "High risk" | – | (1.1-2.6) | (1.3-3.0) |
|  |  | ≤ 0.01 | ≤ 0.001 |
|  |  | Spearman *r* coefficient | |
|  |  | *P*-value | |
| High easTILs | 0.61[b] | – | – |
|  | 0.0001 |  |  |
|  |  | Concordance index ("c-index") | |
| Predicted risk score (CNN output) vs. actual time-to-event data | – | 0.57 | 0.59 |

95% CI: 95% confidence interval. eaTILs (mm2): Density of TILs over tumor region. easTILs: Density of TILs over stroma area that mimics the international TIL working group variable as read by pathologists. eTILs%: Proportion of TILs over tumor cells. esTILs%: Proportion of TILs over stromal cells. etTILs%: Proportion of TILs over all detected cells. sTIL: Stromal TILs. TILs: Tumor-infiltrating lymphocytes.

[a] Outcomes predicted: Distant disease-free survival in patients with higher TILs
scores (Bai et al. (2021)) and distant disease-free survival in patients predicted as "High risk" by ML models (by Bychkov et al. (2022)).

[b] Good correlation found when the CNN11-derived easTLs variable score was
compared with the pathologist-read sTILs assessment.

*E.1. Results of internal validation of the classification and prognostic model*

Wang et al. achieved an AUC that ranged from 0.919 (95% CI 0.884–0.955) to 0.937 (95% CI 0.887–0.987) in different datasets used for internal validation when separating Nottingham Histological Grade 1 and Nottingham Histological Grade 3 invasive breast carcinomas. In addition, those predicted as high grade showed an increased risk for recurrence, with a HR of 2.94 (95% CI 1.24–6.97, $P = .015$). Pathologists' opinions were also used as the ground truth.[41]

*R. Gonzalez et al.*                                                                                                          Journal of Pathology Informatics 15 (2024) 100348
82. Shankar S, Herman B, Parameswaran AG. Rethinking Streaming Machine Learning Evaluation. Published online May 23, 2022. 10.48550/arXiv.2205.11473
83. Symeonidis G, Nerantzis E, Kazakis A, Papakostas GA. MLOps - definitions, tools and challenges. 2022 IEEE 12th Annual Computing and Communication Workshop and Conference (CCWC); 2022. p. 0453–0460. https://doi.org/10.1109/CCWC54503.2022.9720902.
84. Liu G, Xu T, Ma X, Wang C. Your model trains on my data? Protecting intellectual property of training data via membership fingerprint authentication. IEEE Trans Inf Forensics Secur 2022;17:1024–1037. https://doi.org/10.1109/TIFS.2022.3155921.
85. WHO Blue Books Web Site Launched – IARC. IARC News. Accessed February 13, 2023. https://www.iarc.who.int/news-events/who-blue-books-web-site-launched/
86. Cserni G. Histological type and typing of breast carcinomas and the WHO classification changes over time. Pathologica 2020;112(1):25–41. https://doi.org/10.32074/1591-951X-1-20.
87. Sandbank J, Bataillon G, Nudelman A, et al. Validation and real-world clinical application of an artificial intelligence algorithm for breast cancer detection in biopsies. Npj Breast Cancer 2022;8(1):1-11. https://doi.org/10.1038/s41523-022-00496-w.
88. Chan RC, To CKC, Cheng KCT, Yoshikazu T, Yan LLA, Tse GM. Artificial intelligence in breast cancer histopathology. Histopathology 2023;82(1):198–210. https://doi.org/10.1111/his.14820.
89. MEDLINE® 2023 Database Guide. Ovid Database Guide. Published 2023. Accessed September 17, 2023. https://ospguides.ovid.com/OSPguides/medline.htm.
90. Hoeppner MA. NCBI bookshelf: books and documents in life sciences and health care. Nucleic Acids Res 2013;41(Database issue):D1251–D1260. https://doi.org/10.1093/nar/gks1279.
91. Embase: Excerpta Medica Database Guide. Ovid Database Guide. Published 2023. Accessed September 17, 2023. https://ospguides.ovid.com/OSPguides/embase.htm.
92. CINAHL Database | EBSCO. Accessed September 17, 2023. https://www.ebsco.com/products/research-databases/cinahl-database.
93. About IEEE Xplore. Accessed September 17, 2023. https://ieeexplore.ieee.org/Xplorehelp/overview-of-ieee-xplore/about-ieee-xplore.
94. MICCAI Society Publications. Accessed September 17, 2023. http://www.miccai.org/publications/.
95. MICCAI Society Events. Accessed September 17, 2023: http://www.miccai.org/events/.
96. Proceedings on SPIE Digital Library. Accessed September 17, 2023. https://www.spiedigitallibrary.org/conference-proceedings-of-spie?webSyncID=9441a392-2138-922d-be78-38e4ea5e2c5a&sessionGUID=a00d5892-1dbb-6bd0-83b3-094c0cd11686&_ga=2.19419128.1587339309.1694992598-2003971750.1694992598&cm_mc_uid=01701106380816949925985&cm_mc_sid_50300000=65096671694996533306&SSO=1.
18